\documentclass{article}

\PassOptionsToPackage{sort,square,numbers}{natbib}

    \usepackage[preprint]{neurips_2023}

\usepackage[utf8]{inputenc} 
\usepackage[T1]{fontenc}    
\usepackage{hyperref}       
\usepackage{url}            
\usepackage{booktabs}       
\usepackage{amsfonts}       
\usepackage{nicefrac}       
\usepackage{microtype}      

\usepackage{graphicx}
\usepackage{amsmath}
\usepackage{amssymb}
\usepackage{duckuments}
\usepackage{multirow}
\usepackage{tabularx,booktabs}
\usepackage[dvipsnames,table]{xcolor}
\usepackage{pifont}
\usepackage{subfigure}
\usepackage{algorithm}
\usepackage{algorithmic}
\usepackage{wrapfig}
\usepackage{enumitem}

\title{Prompt-Tuning Decision Transformer\\ with Preference Ranking}

\author{
Shengchao Hu\textsuperscript{\rm 1}
\quad 
Li Shen\textsuperscript{\rm 2}\textsuperscript{\rm}\thanks{Corresponding author: Li Shen}
\quad 
Ya Zhang\textsuperscript{\rm 1,3}
\quad 
Dacheng Tao\textsuperscript{\rm 4}
\\
\textsuperscript{\rm 1}Shanghai Jiaotong University, China;
\textsuperscript{\rm 2}JD Explore Academy, China\\
\textsuperscript{\rm 3} Shanghai AI Laboratory, China;
\textsuperscript{\rm 4} The University of Sydney, Australia\\
{\tt\small \{charles-hu,ya\_zhang\}@sjtu.edu.cn; \quad 
\{mathshenli,dacheng.tao\}@gmail.com
}
}

\begin{document}

\maketitle

\begin{abstract}
  Prompt-tuning has emerged as a promising method for adapting pre-trained models to downstream tasks or aligning with human preferences. 
  %
  %
  Prompt learning is widely used in NLP but has limited applicability to RL due to the complex physical meaning and environment-specific information contained within RL prompts. 
  These factors require supervised learning to imitate the demonstrations and may result in a loss of meaning after learning. 
  Additionally, directly extending prompt-tuning approaches to RL is challenging because RL prompts guide agent behavior based on environmental modeling and analysis, rather than filling in missing information, making it unlikely that adjustments to the prompt format for downstream tasks, as in NLP, can yield significant improvements.
  In this work, we propose the Prompt-Tuning DT algorithm to address these challenges by using trajectory segments as prompts to guide RL agents in acquiring environmental information and optimizing prompts via black-box tuning to enhance their ability to contain more relevant information, thereby enabling agents to make better decisions.
  %
  Our approach involves randomly sampling a Gaussian distribution to fine-tune the elements of the prompt trajectory and using preference ranking function to find the optimization direction, thereby providing more informative prompts and guiding the agent towards specific preferences in the target environment.
  Extensive experiments show that with only 0.03\% of the parameters learned, Prompt-Tuning DT achieves comparable or even better performance than full-model fine-tuning in low-data scenarios.
  Our work contributes to the advancement of prompt-tuning approaches in RL, providing a promising direction for optimizing large RL agents for specific preference tasks.

  
\end{abstract}

\section{Introduction}

Pre-trained large-scale models (PLMs) \cite{brown2020language, bert, radford2021learning, chen2020simple} have proven to be highly effective for a wide range of tasks due to their high transferability and competitive performance on downstream tasks with limited data. 
However, full-model fine-tuning requires updating and storing all the parameters of the PLM, which is memory-intensive and impractical for maintaining a separate set of parameters for each task during inference.
Recently, prompt-tuning \cite{li2021prefix, shin2020autoprompt} has emerged as a promising alternative to full-model fine-tuning, allowing for the adaptation of pre-trained models to specific downstream tasks or human preferences.
By freezing the pre-trained model's parameters and tuning only the prompts, prompt-tuning approaches have demonstrated comparable performance with full-model fine-tuning methods across various model scales and tasks \cite{liu2021gpt, lester2021power, zhong2021factual}.

Offline Reinforcement Learning (offline RL) is a data-driven approach that learns an optimal policy from trajectories collected by a set of behavior policies, without requiring access to the environments. 
This approach is critical in many settings, where online interactions are expensive or dangerous. 
However, offline RL struggles with generalization to unseen tasks and adapting to preferences, as the agent may not find a good policy in the test tasks due to the distribution shift. 
Recent works address this challenge through offline meta-RL, which leverages the algorithmic learning perspective \cite{mitchell2021offline, nichol2018first, rajeswaran2019meta}. 
In contrast, we aim to investigate the power of prompt-tuning methods with PLMs. 
Nonetheless, unlike natural language processing (NLP), RL prompts are more complex and contain environment-specific information, which makes it challenging to apply prompt learning to RL. 
Additionally, prompt-tuning approaches from NLP cannot be directly applied to RL prompts, as RL prompts guide agent behavior based on environmental modeling and analysis rather than filling in missing information. 
%
Therefore, there is an urgent need to develop novel prompt-tuning techniques specifically tailored to RL that can guide agents towards specific preferences in the target environment.

In this paper, we propose a new algorithm called Prompt-Tuning Decision Transformer (Prompt-Tuning DT) to address the challenge of generalization in offline RL. 
Our approach leverages trajectory segments as prompts to guide RL agents in acquiring environmental information and optimizes prompts via black-box tuning to enhance their ability to contain more meaningful information.
%
Prompt-tuning is essential in RL as it enables agents to make better decisions by providing more informative prompts that guide the agent towards specific preference in the target environment, which tackles the issue of insufficient knowledge of a PLM to adapt to different preferences and the question of whether a pre-collected prompt can provide enough information about the target tasks.

In our prompt-tuning offline RL framework, we first pre-train the agent using offline trajectories from various tasks within the same environment.
For each task, the agent learns to predict a target trajectory while conditioning on the trajectory prompt sampled from the same task. 
During the evaluation, the agent is presented with a new task and a small set of new trajectories and prompts (total step length not exceeding 10) for fine-tuning the prompt.
%
Our approach randomly samples a Gaussian distribution to perturb each element of the prompt to avoid catastrophic deviations and employs a preference ranking function to determine the optimization direction using a ranking algorithm.
Remarkably, optimizing only 0.03\% of the model parameters, Prompt-Tuning DT achieves comparable performance in the full-data setting and outperforms full-model fine-tuning in the low-data  scenarios.
Our work contributes to the advancement of prompt-tuning approaches in RL, providing a promising direction for optimizing large pre-trained RL agents for specific preferences and downstream tasks.

In summary, our main contributions are three-fold:
\begin{itemize}[leftmargin=*]
  \item We propose Prompt-Tuning DT, a memory-efficient alternative to fine-tuning PLMs that achieves comparable performance with full-model fine-tuning in downstream tasks. 
  
  \item  We present a prompt-tuning offline RL framework using trajectory prompts, enabling effective adaptation to new tasks with minimal parameter optimization and a handful of trajectories.

  \item  By optimizing with preference ranking, Prompt-Tuning DT outperforms strong meta offline RL baselines, demonstrating its effectiveness as a few-shot learner for generalization in offline RL.
\end{itemize}

\begin{figure}
    \centering
    \includegraphics[width=1.0\textwidth]{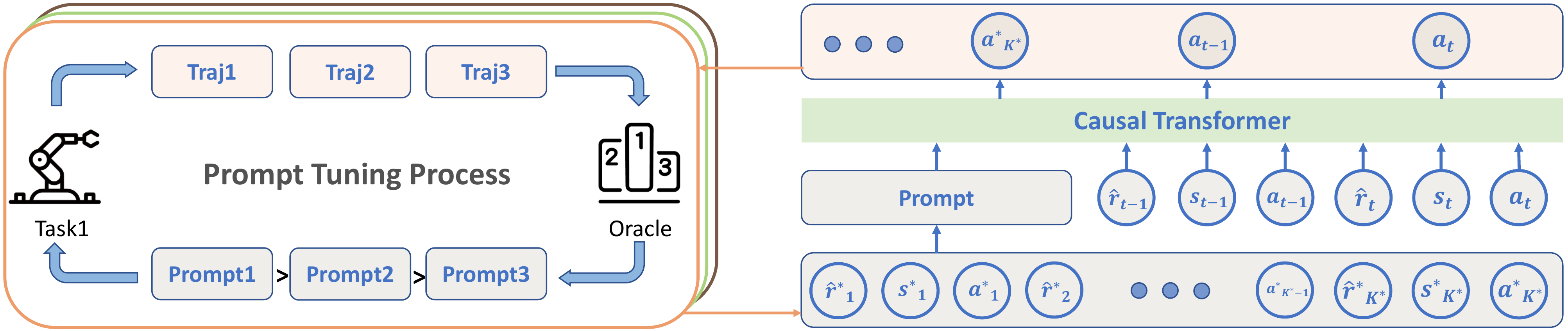}
    \vspace{-0.4cm}
    \caption{Application of Prompt-Tuning DT on few-shot learning setting. At each iteration, the causal transformer generates different trajectories for the task based on different prompts, which are generated by perturbing the initial prompt with random noise, and the most recent K-step history. The generated trajectories are then ranked based on a specific property using a preference ranking function oracle, and the ranking information is leveraged to update the prompt.}
    \label{fig:PTDT}
    \vspace{-0.4cm}
\end{figure}

\section{Related Work}


\textbf{Offline RL.}
Offline RL has emerged as a promising paradigm for learning from fixed, limited datasets consisting of trajectory rollouts from arbitrary policies \cite{levine2020offline}. 
However, deploying off-policy RL algorithms directly in the offline setting can be challenging due to the distributional shift problem, which can result in a significant performance drop \cite{fujimoto2019off}.
To overcome this issue, model-free RL algorithms adopt various strategies, such as constraining the action space of the policy \cite{kumar2019stabilizing, siegel2020keep} or introducing pessimism to the value function \cite{CQL, fujimoto2019off}, to explicitly correct the distributional mismatch between the behavior policy in the offline data and the optimized policy.
In contrast, model-based RL algorithms estimate the reward and transition functions with offline data and require modifications to the RL algorithm to avoid exploiting errors in the estimated model \cite{kidambi2020morel, yu2020mopo, yu2021combo}. 
%

Offline RL has been increasingly viewed as a sequence modeling task, and Transformer-based decision models have been applied to this domain. 
The objective is to predict the next sequence of actions based on the sequence of recent experiences, which includes state-action-reward triplets. 
This approach can be trained using supervised learning, which makes it more suitable for offline RL and imitation learning. 
Several studies have explored the use of Transformers in RL under the sequence modeling paradigm, including QDT \cite{QDT}, Generalized DT \cite{generalizedDT}, Graph DT \cite{hu2023graph}, and the survey works \cite{hu2022transforming,li2023survey}.
In this work, we propose a novel approach that is based on Prompt-DT \cite{PDT} and incorporates prompt-tuning techniques to enhance its performance on downstream target tasks.

\textbf{Meta RL.}
Meta-learning algorithms enable efficient learning of new tasks by learning the process of learning itself.
In the context of meta-RL, the objective is to generalize an agent's knowledge from one task to another.
%
%
%
%
In recent years, several studies have delved into the problem of offline meta-reinforcement learning (meta-RL).
Li et al. \cite{li2020multi} address a scenario where task identity is spuriously inferred due to biased datasets and apply the triplet loss to robustify the task inference with reward relabelling. 
Dorfman et al. \cite{dorfman2021offline} extend the online Meta-RL method VariBAD \cite{zintgraf2019varibad} to the offline setup, where they assume known reward functions for each task and use reward relabelling to share data across tasks with shared dynamics.
On the other hand, Mitchell et al. \cite{mitchell2021offline} propose an offline Meta-RL algorithm based on MAML \cite{finn2017model}. 
Their approach includes an advantage weighting loss and introduces a carefully chosen policy update in the inner loop to theoretically increase the richness of the policy's update and empirically improve adaptation performance and stability.
In this article, we investigate an alternative perspective on meta-RL using sequence modeling and prompt engineering, which can achieve comparable or superior performance compared to traditional methods.

\textbf{Prompt Learning.}
Prompt learning is a promising methodology in NLP that involves optimizing a small subset of parameters while leaving the main model architecture unchanged.
%
%
The basic premise of prompt learning involves presenting the model with a cloze test-style textual prompt, which the model is then expected to fill in with the corresponding answer.
%
%
Autoprompt \cite{shin2020autoprompt} proposes an automatic prompt search methodology for efficiently finding optimal prompts, while recent advancements in prompt learning have adopted trainable continuous embeddings for prompt representation \cite{li2021prefix, lester2021power}.
%
%
Prompt learning has also been applied to the vision-language domain, where introducing continuous prompts into pre-trained vision-language models has led to significant improvements in few-shot visual recognition and generalization performance \cite{zhou2022learning, zhou2022conditional}.
While prompt learning reduces the number of tunable parameters, back-propagation through the entire model is still necessary to calculate gradients and update the small subset of parameters. 
Gradient-free methods have been proposed to optimize continuous \cite{sun2022black, diao2022black} or discrete \cite{prasad2022grips, deng2022rlprompt} prompts. 
Despite the great success of prompt-tuning in the fields of NLP and CV, its application in RL has not been thoroughly explored.
Therefore, in this study, we propose the Promp-Tuning DT method that employs gradient-free methods to optimize continuous trajectory prompts with a ranking oracle, which can be extended to incorporate human preferences by ranking candidate prompts manually.

\section{Preliminary}

In this section, we provide a concise overview of the key components of our algorithm, namely Prompt-DT and ranking optimization. 
Prompt-DT extends the transformer model to offline RL by utilizing trajectory prompts, enabling few-shot generalization to unseen tasks. 
We also introduce a ranking optimization approach that utilizes ranking data to optimize the model without explicit gradient information and present the fundamental rank-based gradient estimator.
These algorithms form the basis of our Prompt-Tuning DT approach that is illustrated in Section \ref{sec:PTDT}.

\subsection{Prompt Decision Transformer}
Transformer \cite{attention}, extensively studied in NLP \cite{bert} and CV \cite{ViT}, has also been explored in RL using the sequence modeling pattern \cite{TRL}. 
Moreover, Recent works from NLP suggest Transformers pre-trained on large-scale datasets are capable of few-shot or zero-shot learning under the prompt-based framework \cite{liu2023pre, brown2020language}. 
%
Inherit from it, Prompt-DT \cite{PDT} for offline RL also leverages the prompt-based framework and adapts it to the context of offline RL to enable few-shot generalization to unseen tasks.
Unlike NLP tasks, where text prompts containing task-specific instructions are used to generate answers without changing the model parameters, Prompt-DT defines a trajectory prompt as a sequence of a few trajectory segments.
Each trajectory segment contains multiple tuples of state $s^*$, action $a^*$ and reward-to-go $\hat{r}^*$, ($s^*$, $a^*$, $\hat{r}^*$), following the trajectory representation in Decision Transformer. 
Each element with superscript $\cdot^*$ is associated with the trajectory prompt.
The reward-to-go is the cumulative reward from the current time step till the end of the episode.
%
Note that the trajectory prompt is typically shorter than the task's horizon, containing only the necessary information to aid in task identification while being insufficient for the agent to imitate.
When training with offline collected data, for each task $\mathcal{T}_i$, Prompt-DT takes $\tau^{input}=(\tau_i^*, \tau_i)$ as input, which contains both the $K^*$ step trajectory prompt and the most recent $K$ step history and can be formulated as 
\begin{equation}
    \tau^{input} = (\hat{r}^*_1, s^*_1, a^*_1, \dots,  \hat{r}^*_{K^*}, s^*_{K^*}, a^*_{K^*}, \hat{r}_1, s_1, a_1, \dots,  \hat{r}_{K}, s_{K}, a_{K}).
\end{equation}
Each head corresponding to a state token is trained to predict an action by minimizing mean-squared loss in the case of continuous action spaces:
\begin{equation}
   L_{DT} = \mathbb{E}_{\tau^{input} \sim \mathcal{T}} \left[ \frac{1}{T} \sum_{t=1}^T (a_t - \pi(\tau^*, \tau_t ) )^2 \right].
\end{equation}

\subsection{Ranking Optimization}
\label{subsec:RO}
Black-box optimization, which utilizes a derivative-free framework to optimize the target function, has been extensively studied in the optimization literature for several decades. 
With the rapid development of Reinforcement Learning with Human Feedback (RLHF), ranking data, which enables humans to express their personal preferences in a straightforward and intuitive manner \cite{bai2022training, ouyang2022training}, has demonstrated great potential for use in various applications, especially those where the exact value of personal information is sensitive, such as healthcare or finance.
ZO-RankSGD \cite{cai2022one,bergou2020stochastic,tang2023zeroth} is an effective approach for model optimization that finds the descent direction directly from the ranking information, without the need for knowledge of the gradient of the model or the exact value of the data.
Given a function $f: \mathbb{R}^d \rightarrow \mathbb{R}$ and $m$ points $x_1, \dots, x_m$ to query, an $(m,k)$ ranking oracle $O_f^{(m,k)}$ will return $k$ smallest points sorted in their order.
With the ranking oracle $O_f^{(m,k)}$ and a starting point $x$, we can query $O_f^{(m,k)}$ with the inputs $x_i = x + \mu \xi_i, ~~\xi_i \sim \mathcal{N}(0, I_d)$, for $i=1,\dots, m$, and $\mu$ is a constant.
With the directed acyclic graph (DAG) $\mathcal{G} = (\mathcal{N}, \mathcal{E})$ constructed from the ranking information of $O_f^{(m,k)}$, where the node set $\mathcal{N} = \{1, \dots, m\}$ and the directed edge set $\mathcal{E} = \{ (i,j)~ |~ f(x_i) < f(x_j) \}$, the rank-based gradient estimator can be formulated as: 
\begin{equation}
    \tilde{g}(x) = \frac{1}{|\mathcal{E}|} \sum_{(i,j) \in \mathcal{E}} \frac{x_j - x_i}{\mu} = \frac{1}{|\mathcal{E}|} \sum_{(i,j) \in \mathcal{E}} (\xi_j - \xi_i).
\end{equation}
Then the point can be updated to $x = x - \eta \tilde{g}(x)$, where $\eta$ is the learning rate.
With the help of preference ranking oracle and ZO-RankSGD  algorithm, we are able to optimize the prompt guiding the agent towards human preference in the target environment.

\section{Prompt-Tuning Decision Transformer}
\label{sec:PTDT}
This section introduces prompt-tuning as a potential alternative to full-model fine-tuning in the context of few-shot policy generalization tasks. 
We begin by presenting the problem formulation in Section \ref{subsec:PF} and subsequently provide a formal definition of our method in Section \ref{subsec_DBBT}.
The overall procedure of our proposed Prompt-Tuning DT is depicted in Figure \ref{fig:PTDT}.

\subsection{Problem Formulation}
\label{subsec:PF}
We consider human preference as aligning with the goal of maximizing rewards and conduct experiments on few-shot policy generalization tasks, which involve training on a set of tasks using offline data and then evaluating the agent's ability to generalize to new tasks.
There are two distinct sets of tasks, denoted as $\mathcal{T}^{train}$ and $\mathcal{T}^{test}$, ensuring that there is no overlap between them ($\mathcal{T}^{train} \cap \mathcal{T}^{test} = \emptyset$). 
This arrangement requires the model to perform well on tasks with goals that lie outside the training range, thereby necessitating the ability to generalize to out-of-distribution tasks.
Each task $\mathcal{T}_i$ in the training set $\mathcal{T}^{train}$ is associated with a corresponding dataset $\mathcal{D}_i$, which consists of pre-collected trajectories obtained using an unknown behavior policy $\pi_i$.
For a test task $\mathcal{T}_i \in \mathcal{T}^{test}$, there are two possible approaches to adapt to the new domain. 
%
One approach involves updating the model parameters using task-specific offline data $\mathcal{P}_i$, which is usually much less than the training dataset $|\mathcal{P}_i| << |\mathcal{D}_i|$.
Alternatively, one can choose to incorporate task information from the prompt, which is constructed from $\mathcal{P}_i$, to mitigate the issue of catastrophic deviations caused by the distribution shift.
%
Our method combines the advantages of both approaches to fine-tune the prompts.

We also introduce a new problem that involves fine-tuning pre-trained policies using limited expert data.
Obtaining expert data often requires human involvement, leading to a restriction on its size.
Conversely, a pre-trained model can be obtained through random algorithms and datasets, which do not necessarily guarantee optimality. 
Therefore, the problem of optimizing a pre-trained agent using limited expert data is an essential area of research that can significantly reduce the cost of human intervention.
Formally, for a given task $\mathcal{T}_i \in \mathcal{T}^{train} = \mathcal{T}^{test}$, the pre-trained agent is provided with an expert dataset $\mathcal{P}_i$ to further fine-tune its policy to maximize the numerical episodic reward signal.
\begin{algorithm}[htbp]
    \small
        \begin{algorithmic}[1]
          \REQUIRE Initial prompt $\tau_0^*$, stepsize $\eta$, iterations $T$, smoothing parameter $\mu$.
        \STATE Construct the initial point from prompt: $x_0 = \hat{r}^*_1 ~||~ s_1^* ~||~ a_1^* ~||~ \dots ~||~ \hat{r}_{K^*}^* ~||~ s_{K^*}^* ~||~ a_{K^*}^*$.
        \FOR{$t=1$ to $T$}
        \STATE Sample $m$ i.i.d. random vectors $\{\xi_{(t,1)},\cdots,\xi_{(t,m)}\}$  from $N(0, I_{d_x})$.
        \STATE Query the ranking function to obtain the exact value with offline loss function \ref{eq:offline} or online reward function \ref{eq:online} with input $\{x_{t-1}+\mu\xi_{(t,1)},\cdots,x_{t-1}+\mu\xi_{(t,m)}\}$. 
        \STATE Construct the corresponding DAG $\mathcal{G}=(\mathcal{N},\mathcal{E})$ as described in Section \ref{subsec:RO}.
        \STATE Compute the gradient estimator  using:
       $g_t = \frac{1}{|\mathcal{E}|} \sum_{(i,j)\in \mathcal{E}}(\xi_{(t,j)}-\xi_{(t,i)})$.
        \STATE $x_t = x_{t-1} - \eta g_t$.
        \ENDFOR
        \end{algorithmic}
        \caption{Prompt-Tuning DT}
        \label{alg:PTDT}
\end{algorithm}

\subsection{Deep Black-Box Tuning}
\label{subsec_DBBT}

In contrast to the prompt learning approach typically employed in NLP, where a cloze test-style textual prompt is presented to the model for filling in the corresponding answer, the trajectory prompt utilized in the decision transformer comprises tokens that possess distinct expressions and physical meanings.
These tokens are designed to represent crucial elements of RL-based tasks, including the state, action, and return-to-go.
The state token encapsulates the environmental information of the agent at a given position and is usually represented by a high-dimensional vector. 
On the other hand, the action token exhibits significant variations across dimensions, with specific values corresponding to distinct actions. 
Moreover, the return-to-go token serves to denote the expected reward that we aim for the agent to attain.
Given these distinct characteristics of RL prompts, directly applying prompt-tuning approaches from NLP becomes challenging: 
RL prompts are specifically tailored to guide agent behavior by leveraging environmental modeling and analysis, rather than primarily focusing on completing missing information as in NLP prompt learning.

We use the ZO-RankSGD to optimize the trajectory prompt without trainable parameters.
Given the initial trajectory prompt $\tau_i^*$, we concatenate one trajectory segment as a unit and then add standard Gaussian distribution to it avoiding catastrophic deviations:
\begin{equation}
    \begin{split}
        x_0 &= \hat{r}^*_1 ~||~ s_1^* ~||~ a_1^* ~||~ \dots ~||~ \hat{r}_{K^*}^* ~||~ s_{K^*}^* ~||~ a_{K^*}^*, \\
        x_i &= x_{i-1} + \mu \xi_i, ~~ \xi_i \sim \mathcal{N}(0, I_{d_x}),
    \end{split}
\end{equation}
where $||$ means concatenation, $\hat{r}^*_i \in \mathbb{R}^{d_r}, s^*_i \in \mathbb{R}^{d_s}, a_i^* \in \mathbb{R}^{d_a}$, and $d_x = (d_r + d_s + d_a) \times K^*$.
For the ranking function $f$, we consider two approaches: offline loss function and online reward function.
Offline loss function adopts the corresponding environmental data $\mathcal{P}_i$ collected in advance, and the loss function value is output according to the normal training process, which can be formulated as:
\begin{equation}
\label{eq:offline}
     f(x_i) = \mathbb{E}_{\tau^{\text{offline}} \sim \mathcal{P}_i} \left[ \frac{1}{T} \sum_{t=1}^T (a_t - \pi(x_i, \tau^{\text{offline}}_t ) )^2 \right].
\end{equation}
While for the online reward function, we take the cumulative reward value obtained by the model in the online environment as output, which is represented as follows:
\begin{equation}
\label{eq:online}
    f(x_i) = -\mathbb{E}_{\tau^{\text{online}} \sim \pi(x_i)} \left[ \sum_{t=1}^{\infty} r_t \right].
\end{equation}
Note that the function is optimized to the minimal, we need to add a minus sign in front of the equation \ref{eq:online}.
Then the ranking oracle $O_f^{(m,k)}$ simply returns the order of these values.
Humans can also perform the oracle online, and the function is not necessarily required to produce an exact value.
Here we only show the effectiveness of the algorithm, and in the following experiments we verify that we only need a small number of oracle calls to approximate the performance of full-model fine-tuning.

We summarize the entire procedure of prompt-tuning in Algorithm \ref{alg:PTDT}. 
Prompt-Tuning DT employs an approximate gradient calculation to adapt the pre-trained agent to specific preferences. 
Gaussian noise is introduced to the initial prompt, driving Prompt-Tuning DT to discover a more expressive prompt tailored to the target tasks.
There are two options available for the ranking function. The offline loss function requires pre-collected datasets from the target tasks in $\mathcal{T}^{test}$, while the online reward function assumes interaction with a simulator for the target tasks in $\mathcal{T}^{test}$.
After $T$ iterations of the fine-tuning process, we utilize the optimized result to initialize the prompt $\tau^*$ at the onset of the evaluation stage and update the recent history $\tau$ with streamingly collected data.
\section{Experiment}
\label{sec:exp}
We perform experiments to assess the aligning human preference of Prompt-Tuning DT by using the episode accumulated reward as the evaluation metric.
%
Our experimental evaluation seeks to answer the following research questions:
(1) Can the prompt-tuning approach achieve comparable performance to full-model fine-tuning with limited ranking oracle calls?
(2) What is the impact of the fine-tuning dataset size on the effectiveness of the prompt-tuning approach?
(3) How does the quality and quantity of the prompt influence the effectiveness of the prompt-tuning approach?

\subsection{Datasets and Tasks}
In this study, we assess the performance of our proposed approach on several datasets that are used in Prompt-DT \cite{finn2017model, rothfuss2018promp, mitchell2021offline, yu2020meta}, namely Cheetah-dir, Cheetah-vel, Ant-dir, and Meta-World reach-v2.
%
%
The objectives of these tasks are distinct, with Cheetah-dir and Ant-dir incentivizing high velocity in the goal direction, Cheetah-vel penalizing deviations from the target velocity using $l_2$ errors, and Meta-World reach-v2 task requiring the robot's end-effector to reach a designated position in 3D space.
In addition, we extend our evaluation to three continuous control tasks, namely HalfCheetah, Hopper, and Walker, selected from the D4RL benchmark \cite{d4rl}, using two different offline dataset settings: Medium and Expert. 
These tasks require the agent to move forward as far and quickly as possible.
%
%
Detailed environment and hyper-parameter settings refer to the supplementary.

We adopt the dataset construction and settings from \cite{mitchell2021offline} for the meta-RL control tasks considered in this study. 
Specifically, the datasets comprise the full replay buffer of Soft Actor-Critic \cite{SAC} for Cheetah-dir and Ant-dir, and TD3 \cite{TD3} for Cheetah-vel. 
Expert trajectories are collected for Meta-World reach-v2 \cite{yu2020meta} using script expert policies provided in the environment. 
For the OpenAI Gym tasks, the medium datasets are generated by a policy that achieves approximately one-third of the score of an expert policy, while expert datasets are generated directly by an expert policy.

\begin{table*}[tb!]
\centering
\caption{Results for meta-RL control tasks.
%
The best mean scores are highlighted in bold.
Each environment has prompts of length $K^*=5$ and is fine-tuned at the full replay buffer with three distinct seeds.
%
%
Notably, all fine-tuning methods outperform the approaches that do not incorporate additional fine-tuning on the left side of the table, and PTDT-offline achieves the best average result.
}
\label{tab:meta}
\vspace{.2cm}
\scalebox{0.78}{%
\begin{tabular}{>{\centering}p{0.14\textwidth}|>{\centering}p{0.14\textwidth}>{\centering}p{0.18\textwidth}>{\centering}p{0.14\textwidth}|>{\centering}p{0.18\textwidth}>{\centering}p{0.14\textwidth}>{\centering\arraybackslash}p{0.14\textwidth}}
\toprule
Environment &   Prompt-DT & Prompt-MT-BC & MT-ORL & Prompt-DT-FT & PTDT-offline & PTDT-online \\ \midrule
Cheetah-dir  &   934.6 $\pm$ 4.4 & 933.2 $\pm$ 0.8  & -46.2 $\pm$ 3.4 &  936.9 $\pm$ 4.8 &  \textbf{941.5 $\pm$ 3.2} & 940.8 $\pm$ 2.3 \\
Cheetah-vel       &  -43.5 $\pm$ 3.4 & -43.8 $\pm$ 5.2  & -146.6 $\pm$ 2.1 & -40.1 $\pm$ 3.8  & \textbf{-39.5 $\pm$ 3.7} &  \textbf{-39.5 $\pm$ 3.3} \\
Ant-dir       &  420.2 $\pm$ 5.1 & 426.8 $\pm$ 8.0 & 110.5 $\pm$ 2.2  &  425.2 $\pm$ 8.6 & \textbf{427.9 $\pm$ 4.3} & 426.2 $\pm$ 4.7 \\ 
MW reach-v2       &  469.8 $\pm$ 29.9 & 447.3 $\pm$ 15.0 & 264.1 $\pm$ 9.6 &  \textbf{478.1 $\pm$ 27.8} & 472.5 $\pm$ 29.0  & 472.2 $\pm$ 29.3 \\\midrule
\textbf{Average} &445.3 & 440.9 & 181.8  & 450.0 &  \textbf{450.3} & 449.9 \\ \bottomrule
\end{tabular}%
}
\vspace{-0.4cm}
\end{table*}

\begin{table*}[tb!]
\centering
\caption{Results for D4RL with medium datasets. 
%
The best mean scores are highlighted in bold.
Each environment has prompts of length $K^*=5$ and is fine-tuned at the limited expert dataset with three different seeds.
The results reveal that both PTDT-offline and PTDT-online methods exhibit similar performance, and they both achieve comparable results to full-model fine-tuning approaches. 
%
}
\label{tab:d4rl}
\vspace{.2cm}
\scalebox{0.8}{%
\begin{tabular}{>{\centering}p{0.14\textwidth}|>{\centering}p{0.14\textwidth}>{\centering}p{0.14\textwidth}>{\centering}p{0.14\textwidth}|>{\centering}p{0.18\textwidth}>{\centering}p{0.14\textwidth}>{\centering\arraybackslash}p{0.14\textwidth}}
\toprule
Environment & CQL &   DT & Prompt-DT & Prompt-DT-FT  & PTDT-offline & PTDT-online \\ \midrule
HalfCheetah  & \textbf{44.4} & 42.6 $\pm$ 0.1 & 42.5 $\pm$ 0.0  & 42.7 $\pm$ 0.1 &  42.7 $\pm$ 0.1 &  42.6 $\pm$ 0.2 \\
Hopper       & 58.0 &   67.6 $\pm$ 1.0 & 68.9 $\pm$ 1.6  & \textbf{71.1 $\pm$ 1.7} & 70.4 $\pm$ 2.0  & 69.7 $\pm$ 2.3 \\
Walker       & \textbf{79.2} &  74.0 $\pm$ 1.4 & 74.6 $\pm$ 2.7 & 75.7 $\pm$ 1.5 &  76.3 $\pm$ 1.7 & 76.2 $\pm$ 1.3 \\ \midrule
\textbf{Average} & 60.5 &61.6 & 62.0 & \textbf{63.2} &  63.1 & 62.8 \\ \bottomrule
\end{tabular}%
}
\vspace{-0.4cm}
\end{table*}

\subsection{Baselines}

We evaluate the few-shot generalization ability of Prompt-Tuning DT by comparing it with four baselines for meta-RL control tasks.
Three of these are few-shot learners that do not undergo any extra fine-tuning on unseen target tasks, while the fourth is a baseline that employs conventional full-model fine-tuning on the target datasets. 
We consider three offline few-shot learners: Prompt-DT, Multi-task Offline RL (MT-ORL), and Prompt-based Behavior Cloning (Prompt-MT-BC). 
Prompt-DT is the first application of sequence-prediction models that achieve state-of-the-art offline meta-RL algorithms in multiple benchmark domains.
MT-ORL omits the prompt augmentation step used in Prompt-DT to construct a variant of the approach. 
Prompt-MT-BC removes the reward-to-go tokens stored in the history input of Prompt-DT and achieves better performance than MACAW \cite{mitchell2021offline} in all environments by a large margin, which is the conventional state-of-the-art offline meta-RL algorithm.
After training the Prompt-DT model, we perform an additional full-model fine-tuning process with gradient steps using the data collected from the target task, which is referred to as Prompt-DT-FT.

In the D4RL benchmark, we conduct a comprehensive evaluation of our approach, comparing it against three baseline methods that do not utilize fine-tuning. 
These baselines include CQL \cite{CQL}, a state-of-the-art model-free technique for offline RL, as well as DT \cite{DT} and Prompt-DT, which are trained using medium datasets and expert prompts.
Furthermore, we introduce Prompt-DT-FT, which incorporates an additional full-model fine-tuning process on limited expert datasets that is less than 256 samples.
To ensure fair comparisons, we adhere to the normalization protocol proposed in \cite{hafner2020mastering}, wherein the final scores are normalized such that a score of 100 represents expert-level performance, while a score of 0 corresponds to the performance of a random policy.
Note that the results of CQL and DT come from their original paper, while the remaining results are obtained by us.

It is worth noting that the results obtained using our prompt-tuning method rely on the pre-trained Prompt-DT agent.
Our approach consists of two variants: Prompt-Tuning DT with offline loss function (PTDT-offline) and Prompt-Tuning DT with online reward function (PTDT-online). 
To ensure a fair comparison, all offline fine-tuning methods are constrained to access the same amount of samples collected from the target task $\mathcal{P}_i$.
While PTDT-online requires interaction with a simulator, we carefully control the number of interactions to ensure access to new trajectories of similar size.

\subsection{Main Results}
We perform a comparative analysis between Prompt-Tuning DT and the baseline methods to assess their few-shot generalization abilities and evaluate the tuning efficiency of Prompt-Tuning DT in relation to the full-model fine-tuning approach.
%
We use the average episode accumulated reward in the test task set $\mathcal{T}^{test}$ as the evaluation metric. 
The main results are presented in Tables \ref{tab:meta} and \ref{tab:d4rl}, which showcase the few-shot performance of various algorithms.
%
%

\textbf{Meta-RL Control Task.} 
Comparing Prompt-MT-BC and MT-ORL reveals that trajectory prompts contain sufficient information to fully specify the task.
Further comparison between Prompt-DT and Prompt-MT-BC demonstrates that reward-to-go tokens provide additional task-related information, leading to improved results. 
The remaining fine-tuning methods are built upon Prompt-DT, with a uniform size for fine-tuning offline data to ensure a fair comparison.
PTDT-online computes an approximate gradient online, treating episodic units as samples, and restricts the number of samples accessed to ensure equal access to downstream data among all fine-tuning methods.
%
All three fine-tuning methods outperform Prompt-DT, with prompt-tuning achieving comparable or superior performance to full-model fine-tuning while utilizing only approximately 0.03\% of the parameters. 
These results highlight the effectiveness of our proposed prompt-tuning method.

\textbf{D4RL Benchmark.} 
The DT method exclusively trains on the medium dataset, whereas the Prompt-DT method incorporates the expert dataset as a prompt during training.
The prompt length is restricted to 5, encompassing essential information to identify the distribution of the expert dataset while avoiding direct imitation.
With access to a small number of expert prompts, Prompt-DT slightly improves performance compared to DT. 
Furthermore, conducting additional fine-tuning of the entire model on the expert dataset, based on the pre-trained Prompt-DT, significantly improves its performance compared to the baseline. 
In contrast, prompt-tuning achieves comparable performance improvements while utilizing only 0.03\% of the full-model parameters.

\subsection{Ablation}
\label{subsec:ab}
We conduct three ablation studies to investigate specific aspects of our prompt-tuning methods, primarily on the meta-RL control environments.
These ablation studies aim to provide insights into the crucial factors of prompt tuning, including sample efficiency, prompt initialization, and prompt length, which are vital for the effectiveness and practicality of this method.

\begin{table*}[tb!]
\centering
\caption{Ablation on the effect of prompt initialization on the fine-tuning methods. 
We vary the quality of the prompt and datasets in the  Cheetah-vel environment with three levels: Expert, Medium, and Random.
%
Each experiment utilizes prompts of length $K^*=5$, and we assess the performance by conducting fine-tuning on 256 samples with three distinct seeds.
The results highlight the significant impact of prompt initialization on the effectiveness of the prompt-tuning method.
}
\label{tab:abinit}
\vspace{.2cm}
\scalebox{0.8}{%
\begin{tabular}{>{\centering}p{0.16\textwidth}|>{\centering}p{0.14\textwidth}>{\centering}p{0.14\textwidth}>{\centering}p{0.14\textwidth}|>{\centering}p{0.14\textwidth}>{\centering}p{0.14\textwidth}>{\centering\arraybackslash}p{0.14\textwidth}}
\toprule
Prompt & \multicolumn{3}{c|}{Prompt-DT-FT} & \multicolumn{3}{c}{Prompt-Tuning DT offline} \\
 Initialization&
  \multicolumn{1}{c}{Exp. Prompt} &
  \multicolumn{1}{c}{Med. Prompt} &
  \multicolumn{1}{c|}{Ran. Prompt} &
  \multicolumn{1}{c}{Exp. Prompt} &
  \multicolumn{1}{c}{Med. Prompt} &
  \multicolumn{1}{c}{Ran. Prompt} \\ \midrule
Exp. Dataset     & -42.7 $\pm$ 5.9  &  -42.6 $\pm$ 3.7  & -49.7 $\pm$ 11.0 &  -42.1 $\pm$ 1.3   &  -42.7 $\pm$ 6.3 &  -49.9 $\pm$ 1.6   \\
Med. Dataset    & -48.1 $\pm$ 4.2  & -53.9 $\pm$ 2.6 & -50.0 $\pm$ 3.9  & -41.4 $\pm$ 0.8 & -42.9 $\pm$ 5.6  &  -50.3 $\pm$ 1.9  \\
Ran. Dataset    & -51.0 $\pm$ 6.9  & -55.3 $\pm$ 8.2 & -61.4 $\pm$ 3.4   & -41.2 $\pm$ 0.8 &  -42.1 $\pm$ 6.3 &  -50.4 $\pm$ 1.9 \\ \bottomrule
\end{tabular}%
}
\vspace{-0.4cm}
\end{table*}

\begin{table*}[tb!]
\centering
\caption{Ablation on the effect of prompt length on the prompt-tuning method. 
We vary the length of the prompt $K^*$ in four environments.
%
Each environment is fine-tuned using the complete replay buffer with three random seeds.
The results demonstrate that the prompt-tuning method exhibits limited sensitivity to the prompt length and effectively refines the prompt during the fine-tuning process.
}
\label{tab:ablength}
\vspace{.2cm}
\scalebox{0.8}{%
\begin{tabular}{>{\centering}p{0.16\textwidth}|>{\centering}p{0.14\textwidth}>{\centering}p{0.14\textwidth}>{\centering}p{0.14\textwidth}|>{\centering}p{0.14\textwidth}>{\centering}p{0.14\textwidth}>{\centering\arraybackslash}p{0.14\textwidth}}
\toprule
\multirow{2}{*}{Prompt Length} & \multicolumn{3}{c|}{Prompt-DT} & \multicolumn{3}{c}{Prompt-Tuning DT offline} \\
 &
  \multicolumn{1}{c}{$K^*=2$} &
  \multicolumn{1}{c}{$K^*=5$} &
  \multicolumn{1}{c|}{$K^*=10$} &
  \multicolumn{1}{c}{$K^*=2$} &
  \multicolumn{1}{c}{$K^*=5$} &
  \multicolumn{1}{c}{$K^*=10$} \\ \midrule
Cheetah\_dir          & 939.9 $\pm$ 5.8 & 934.6 $\pm$ 4.4       & 930.3 $\pm$ 8.1  & 944.2 $\pm$ 3.2  &   941.5 $\pm$ 3.2  &  941.3 $\pm$ 9.5  \\
cheetah\_vel          &  -45.4 $\pm$ 9.5  & -43.5 $\pm$ 3.4       & -42.1 $\pm$ 3.6 &  -41.5 $\pm$ 9.9   &   -39.5 $\pm$ 3.7  & -40.3 $\pm$ 1.2  \\
ant\_dir              & 413.7 $\pm$ 27.1  &  420.2 $\pm$ 5.1      &  388.8 $\pm$ 2.9 & 422.3 $\pm$ 20.4  &  427.9 $\pm$ 4.3  & 393.4 $\pm$ 5.8 \\
MW reach-v2     & 445.5 $\pm$ 14.8 &  469.8 $\pm$ 29.9  & 462.1 $\pm$ 19.3   & 446.8 $\pm$ 14.7  &  472.5 $\pm$ 29.0 & 462.2 $\pm$ 19.4  \\ \bottomrule
\end{tabular}%
}
\vspace{-0.2cm}
\end{table*}

\begin{figure}
    \centering
    \includegraphics[width=1.0\textwidth]{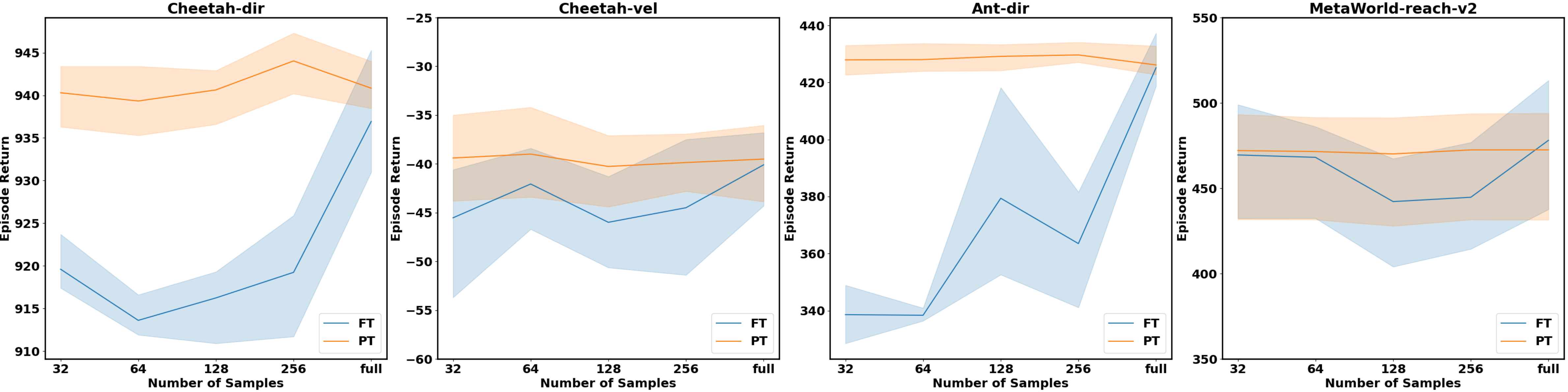}
    \vspace{-0.4cm}
    \caption{Comparison between prompt-tuning (PT, orange) and full-model fine-tuning (FT, blue) when different numbers of training samples are available. 
    The figures from left to right correspond to Cheetah-dir, Cheetah-vel, Ant-dir, and MetaWorld-reach-v2 environments respectively.
    Each plot is run with 3 seeds.
    PT outperforms FT in low-data regimes in addition to requiring many fewer parameters. 
    When the number grows, the performance of the two methods becomes closer.}
    \label{fig:absize}
    \vspace{-0.4cm}
\end{figure}

\textbf{Sample Efficiency.} 
%
%
%
We investigate the effect of increasing the number of fine-tuning samples on the performance of Prompt-DT-FT and PTDT-offline, aiming to understand the prompt-tuning method's dependence on the quantity of fine-tuning samples.
Figure \ref{fig:absize} illustrates the performance trends of these methods on the Cheetah-dir, Cheetah-vel, Ant-dir, and MetaWorld-reach-v2 environments. 
%
PTDT-offline consistently outperforms Prompt-DT-FT with 32 to 256 samples, and their performance gradually becomes closer as the number of samples grows.
When fine-tuning on the full data, PTDT-offline achieves comparable performance with the Prompt-DT-FT.

Furthermore, it is worth noting that unlike the observed phenomenon in NLP \cite{li2021prefix, gu2021ppt}, the performance of the two fine-tuning approaches does not exhibit a monotonically increasing trend as the amount of fine-tuning data increases. 
%
%
%
In all environments, a downward inflection point is observed in the performance curve as the number of samples increases. 
This can be attributed to the presence of "bad samples" in the training dataset, which adversely impact the fine-tuning process and potentially result in catastrophic deviations. 
By utilizing a larger dataset, the proportion of "bad samples" decreases, enabling the fine-tuning to converge to more optimal policies and improve overall performance.

\textbf{Prompt Initialization.}
%
%
Prompt initialization plays a vital role in guiding the agent's behavior and shaping the learning process \cite{gu2021ppt}. 
However, making significant modifications to the prompt during fine-tuning can be challenging when the number of preference ranking oracle querying is limited. 
%
%
%
%
The objective of this ablation study is to understand the importance of prompt initialization and its influence on overall performance. 
To achieve this, we conduct an ablation study in Cheetah-vel and create three datasets - expert, medium, and random - by selecting the last, middle, and first 256 trajectories from the full replay buffer collected in the test environments.
Subsequently, we sample expert, medium, and random few-shot demonstrations from these datasets.
%
The results are shown in Table \ref{tab:abinit}.
Full-model fine-tuning exhibits greater sensitivity to the dataset, whereas the prompt-tuning method demonstrates higher sensitivity to the prompt used.
Initialization with an expert or medium prompt allows the prompt-tuning method to converge to comparable performance. 
However, when initialized with a random prompt, a noticeable performance decline is observed, although it still outperforms the results of full-model fine-tuning in the majority of settings.

\textbf{Prompt Length.}
%
We investigate the impact of prompt length on the prompt-tuning method, considering its influence on both the number of tuning parameters in the approach and the speed of inference.
It is crucial to strike a balance between the richness of information provided by the prompt and the effectiveness of the prompt-tuning process. 
Through this ablation study, we aim to gain insights into identifying the optimal prompt length that ensures satisfactory performance while preserving the effectiveness of the fine-tuning process.
The results are shown in Table \ref{tab:ablength}.
Increasing the amount of information does not significantly improve the generalization performance across all four environments, and Prompt-Tuning DT consistently outperforms Prompt-DT across various prompt lengths, indicating that our approach can effectively fine-tune the prompt towards specific preferences without being overly sensitive to the number of tuning parameters.

\section{Conclusion}

In this paper, we introduce Prompt-Tuning Decision Transformer (Prompt-Tuning DT), a novel algorithm that aligns with human preference in the target environment.
By optimizing prompts without back-propagation, Prompt-Tuning DT offers a memory-efficient alternative to fine-tuning PLMs. 
Furthermore, our prompt-tuning offline RL framework using trajectory prompts allows for effective adaptation to new tasks with minimal parameter optimization and a small number of trajectories.
%
Through extensive experiments, our approach achieves comparable performance to full-model fine-tuning in the full-data setting and outperforms it in the low-data scenarios.
%
We also show that Prompt-Tuning DT is robust to prompt length changes but sensitive to prompt initialization.

\textbf{Broader Impacts.} Our work contributes to the advancement of prompt-tuning approaches in RL, providing a promising direction for optimizing large pre-trained RL agents for specific preferences and downstream tasks.
Our approach demonstrates the potential of prompt-based methods in offline RL settings and opens up avenues for future research in developing tailored prompt-tuning techniques for RL agents.
We envision that prompt-tuning approaches will continue to play a crucial role in enhancing the generalization and adaptability of RL agents in real-world scenarios.

\textbf{Limitation.} We summarize the limitations of this work as follows:
(1) The exploration of the human-in-the-loop environment remains to be undertaken, as it may involve encountering different requirements.
(2) The scalability of the Prompt-Tuning DT method to larger and more complex pre-trained agents and RL environments remains to be explored.

\newpage
{
\small
\bibliographystyle{plainnat}
\bibliography{egbib}
}

\newpage
\appendix
\noindent{\Large \textbf{Appendix}}

We offer a comprehensive description of the environments in Section \ref{sec:apde}, outline the hyperparameters for Prompt-Tuning DT and baselines in Section \ref{sec:aph}, present the ablation studies in Section \ref{sec:apas}, and provide detailed results in Section \ref{sec:apdr}.

\section{Detailed Environment}
\label{sec:apde}
We evaluate our approach on a variety of tasks, including meta-RL control tasks and continuous control tasks. These tasks can be described as follows:
\begin{itemize}[leftmargin=*]
    \item Cheetah-dir: The task comprises two directions: forward and backward, in which the cheetah agent is incentivized to attain high velocity along the designated direction. Both the training and testing sets encompass these two tasks, providing comprehensive coverage of the agent's performance.
    \item Cheetah-vel: In this task, a total of 40 distinct tasks are defined, each characterized by a different goal velocity.  The target velocities are uniformly sampled from the range of 0 to 3. The agent is subjected to a penalty based on the $l_2$ error between its achieved velocity and the target velocity. We reserve 5 tasks for testing purposes and allocate the remaining 35 tasks for training.
    \item Ant-dir: There are 50 tasks in Ant-dir, where the goal directions are uniformly sampled in a 2D space. The 8-joint ant agent is rewarded for achieving high velocity along the specified goal direction. We select 5 tasks for testing and use the remaining tasks for training.
    \item Meta-World reach-v2: This task involves controlling a Sawyer robot's end-effector to reach a target position in 3D space. The agent directly controls the XYZ location of the end-effector, and each task has a different goal position. We train on 15 tasks and test on 5 tasks.
    \item HalfCheetah: The HalfCheetah is a 2-dimensional robot with 9 links and 8 joints. The primary objective is to exert torque on these joints, thereby enabling the cheetah to achieve maximum forward running speed. Positive rewards are assigned based on the distance covered in the forward direction, while negative rewards are incurred for any backward movement. 
    \item Hopper: The Hopper environment features a one-legged figure with four main body parts. The objective of this task is to generate hops in the forward direction through the application of torques on the hinges that connect these body parts. 
    \item Walker2d: In this environment, the goal is to coordinate the movement of a two-legged figure with six body parts. By applying torques to the hinges, the objective is to move in the forward direction.
\end{itemize}

By evaluating our approach on these diverse tasks, we can assess its performance and generalization capabilities across different control scenarios.

The generalization capability of our approach is evaluated by examining the task index of the training and testing sets, as shown in Table \ref{tab:set_meta}. The experimental setup in Section \ref{sec:exp} adheres to the training and testing division specified in Table \ref{tab:set_meta}. This ensures consistency and allows for a comprehensive assessment of the approach's performance across different tasks.

\begin{table}[H]
    \caption{Training and testing task indexes when testing the generalization ability in meta-RL tasks }
    \label{tab:set_meta}
    \centering
    \begin{tabular}{ll}
      \toprule
      \multicolumn{2}{c}{Cheetah-dir} \\
      \midrule
      Training set of size 2 & [0,1] \\
      Testing set of size 2 & [0.1]\\
      \midrule
      \multicolumn{2}{c}{Cheetah-vel} \\
      \midrule
      Training set of size 35 & [0-1,3-6,8-14,16-22,24-25,27-39]\\
      Testing set of size 5 & [2,7,15,23,26]\\
      \midrule
      \multicolumn{2}{c}{ant-dir} \\
      \midrule
      Training set of size 45 &  [0-5,7-16,18-22,24-29,31-40,42-49]\\
      Testing set of size 5 &  [6,17,23,30,41]\\
      \midrule
      \multicolumn{2}{c}{Meta-World reach-v2} \\
      \midrule
      Training set of size 15 &  [1-5,7,8,10-14,17-19]\\
      Testing set of size 5 &  [6,9,15,16,20]\\
      \bottomrule
    \end{tabular}
    \vspace{-0.2in}
\end{table}

\section{Hyperparameters}
\label{sec:aph}
We show the hyperparameters of Prompt-Tuning DT in Table \ref{tab:hp_ptdt} and Prompt-DT in Table \ref{tab:hp_pdt}.

\begin{table}[h]
    \caption{Common Hyperparameters of Prompt-Tuning DT.}
    \label{tab:hp_ptdt}
    \centering
    \begin{tabular}{ll}
      \toprule
      Hyperparameters & Value \\
      \midrule
      $K$ (length of context $\tau$)  & 20 \\
      training batch size for each task   & 32 \\
      training number of steps per iteration & 10 \\
      training max iterations & 5000 \\
      fine-tuning batch size for each task & 32 \\
      fine-tuning number of steps per iteration (offline) & 8 \\
      fine-tuning number of steps per iteration (online) & 10 \\
      fine-tuning max iterations & 20 \\
      number of evaluation episodes for each task  & 50 \\
      learning rate & 1e-4 \\
      learning rate decay weight & 1e-4\\
      ranking algorithm m & 15 \\
      ranking algorithm k & 15 \\
      Return-to-go conditioning & 1500 Cheetah-dir \\
       & 0 Cheetah-vel \\
       & 500 Ant-dir \\
       & 650 Meta-World reach-v2 \\
       & 6000 HalfCheetah \\
       & 3600 Hopper \\
       & 5000 Walker \\
      Pompt length $K^*$ & 5 \\
      number of layers & 3 \\
      number of attention heads & 1 \\
      embedding dimension & 128 \\
      activation & ReLU \\
      \bottomrule
    \end{tabular}
\end{table}

\begin{table}[h]
    \caption{Common Hyperparameters of Prompt-DT, Prompt-MT-BC, and MT-ORL.}
    \label{tab:hp_pdt}
    \centering
    \begin{tabular}{ll}
      \toprule
      Hyperparameters & Value \\
      \midrule
      $K$ (length of context $\tau$)  & 20 \\
      training batch size for each task   & 8 \\
      number of evaluation episodes for each task  & 20 \\
      learning rate & 1e-4 \\
      learning rate decay weight & 1e-4\\
      Return-to-go conditioning & 1500 Cheetah-dir \\
       & 0 Cheetah-vel \\
       & 500 Ant-dir \\
       & 650 Meta-World reach-v2 \\
       & 6000 HalfCheetah \\
       & 3600 Hopper \\
       & 5000 Walker \\
      Pompt length $K^*$ & 5 \\
      number of layers & 3 \\
      number of attention heads & 1 \\
      embedding dimension & 128 \\
      activation & ReLU \\
      \bottomrule
    \end{tabular}
\end{table}

\newpage
\section{Ablation Study}
\label{sec:apas}
In this section, we provide additional visual supplements to enhance the intuitiveness of our experimental results in the ablation study. 
These supplementary visuals aim to provide a clearer representation of our findings and further support our conclusions.

\begin{figure}[h]
    \centering
    \includegraphics[width=1.0\textwidth]{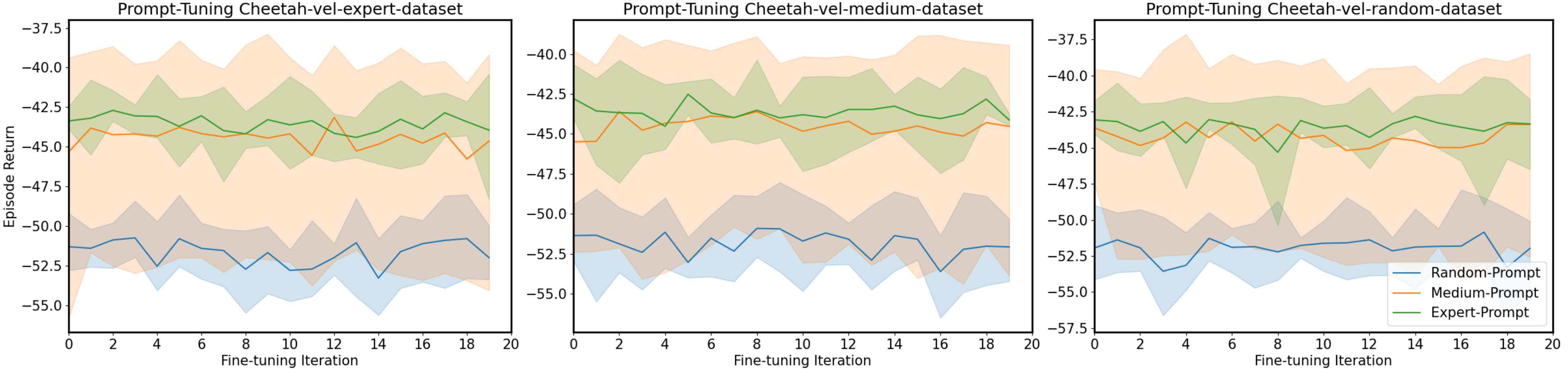}
    \caption{
    Ablation on the effect of prompt initialization on the prompt-tuning method. 
    We vary the quality of the prompt and datasets in the  Cheetah-vel environment with three levels: Expert, Medium, and Random.
    %
    Each experiment utilizes prompts of length $K^*=5$, and we assess the performance by conducting fine-tuning on 256 samples with three distinct seeds.
    The results highlight the significant impact of prompt initialization on the effectiveness of the prompt-tuning method.}
    \label{fig:abinit}
\end{figure}

\begin{figure}[h]
    \centering
    \includegraphics[width=1.0\textwidth]{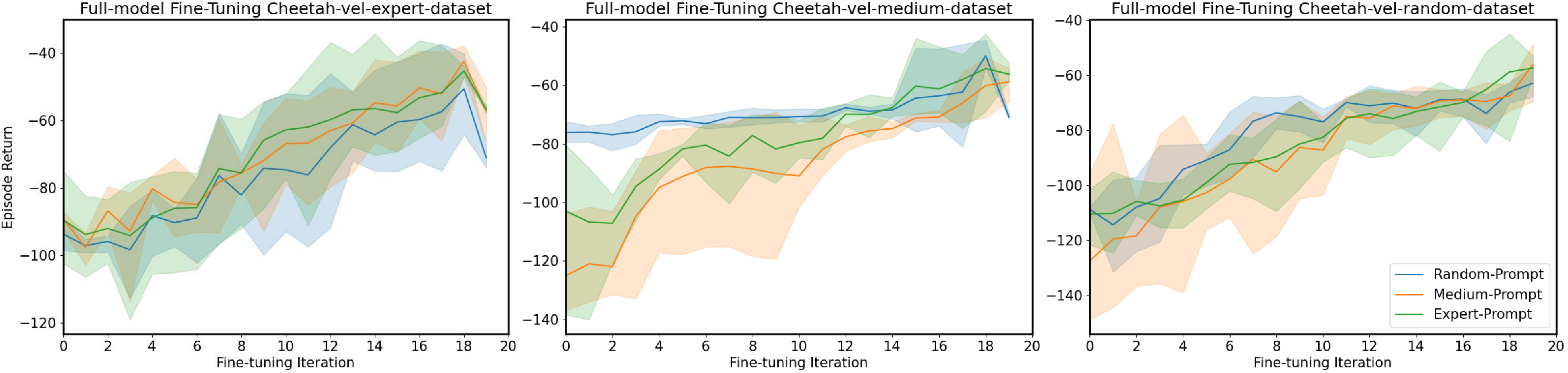}
    \caption{
    Ablation on the effect of prompt initialization on the full-model fine-tuning method. 
    We vary the quality of the prompt and datasets in the  Cheetah-vel environment with three levels: Expert, Medium, and Random.
    %
    Each experiment utilizes prompts of length $K^*=5$, and we assess the performance by conducting fine-tuning on 256 samples with three distinct seeds.
    The results highlight the significant impact of fine-tuning datasets on the effectiveness of the full-model fine-tuning method.}
    \label{fig:abpinit}
\end{figure}

The performance trends during the fine-tuning process are depicted in Figure \ref{fig:abinit} and \ref{fig:abpinit}.
The prompt-tuning method exhibits a notable convergence to high performance from the initial stages, steadily improving with each update. In contrast, the full-model fine-tuning method experiences a catastrophic deviation early on, resulting in performance dropping below -120. While its performance gradually recovers with the progress of fine-tuning, it remains inferior to the prompt-tuning method. 
Furthermore, it is evident that prompt-tuning is more responsive to the quality of prompt initialization, while full-model fine-tuning is more influenced by the quality of the fine-tuning datasets.

\begin{figure}[h]
    \centering
    \includegraphics[width=1.0\textwidth]{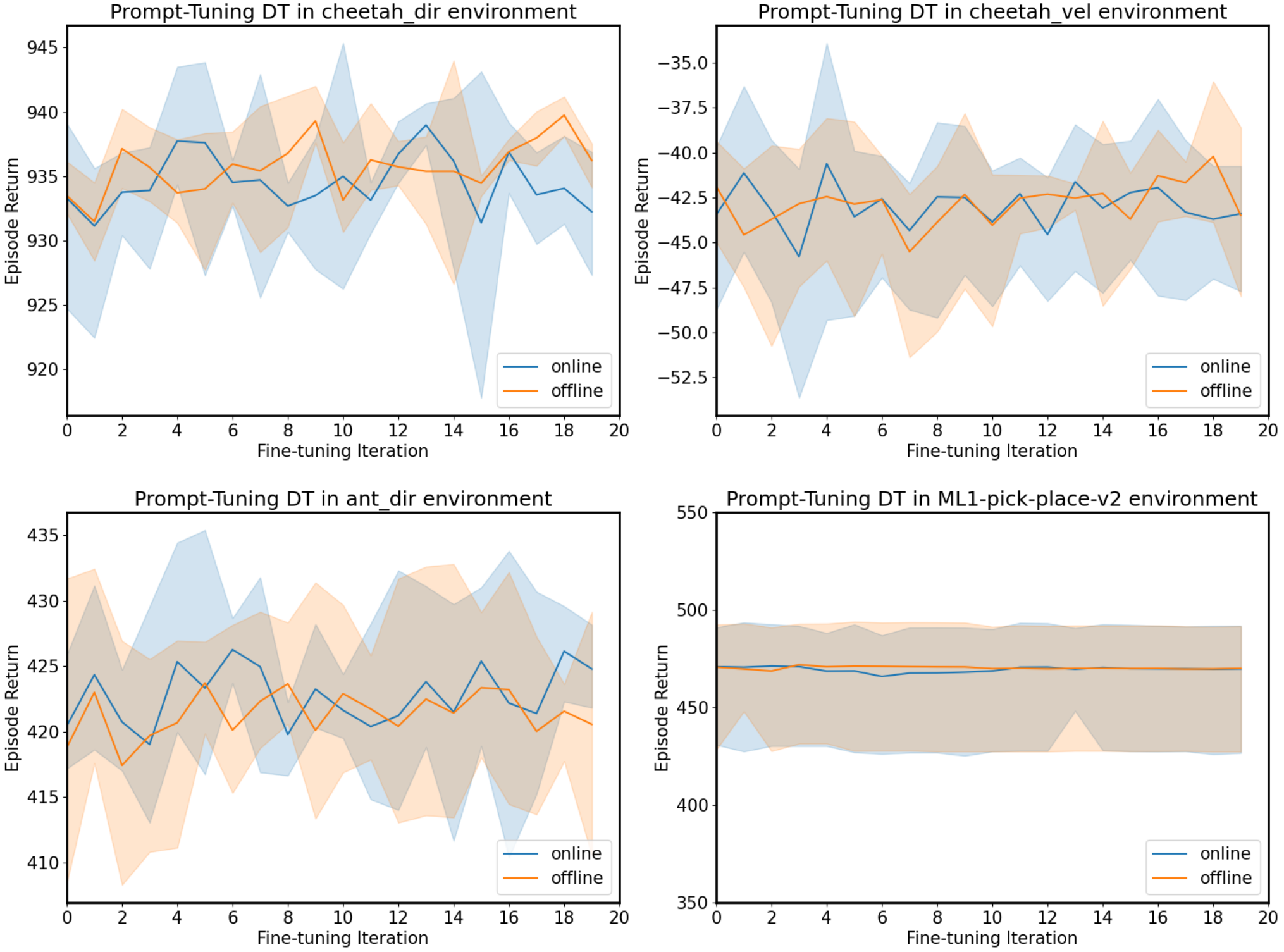}
    \caption{
    Ablation on the comparison between online and offline Prompt-Tuning DT in the four meta-RL control environments.
    We vary the ranking function to offline loss function and online reward function, and evaluate the performance with three different seeds using prompts of length $K^*=5$. 
    The results indicate comparable performance between the online and offline methods, confirming that we have successfully ensured equitable access to target task datasets for both approaches. }
    \label{fig:abstatus}
\end{figure}

We further visualize the fine-tuning process of online and offline Prompt-Tuning DT methods in Figure \ref{fig:abstatus}. 
The results demonstrate a similar performance between these two methods. 
This similarity indicates that we have successfully ensured that both online and offline approaches have access to a comparable number of target task datasets. 
Additionally, the visualization provides insights into the number of human preference ranking oracle queries required for online tuning, which serves as an initial reference for subsequent human-in-the-loop interactions.

\begin{figure}[H]
    \centering
    \includegraphics[width=1.0\textwidth]{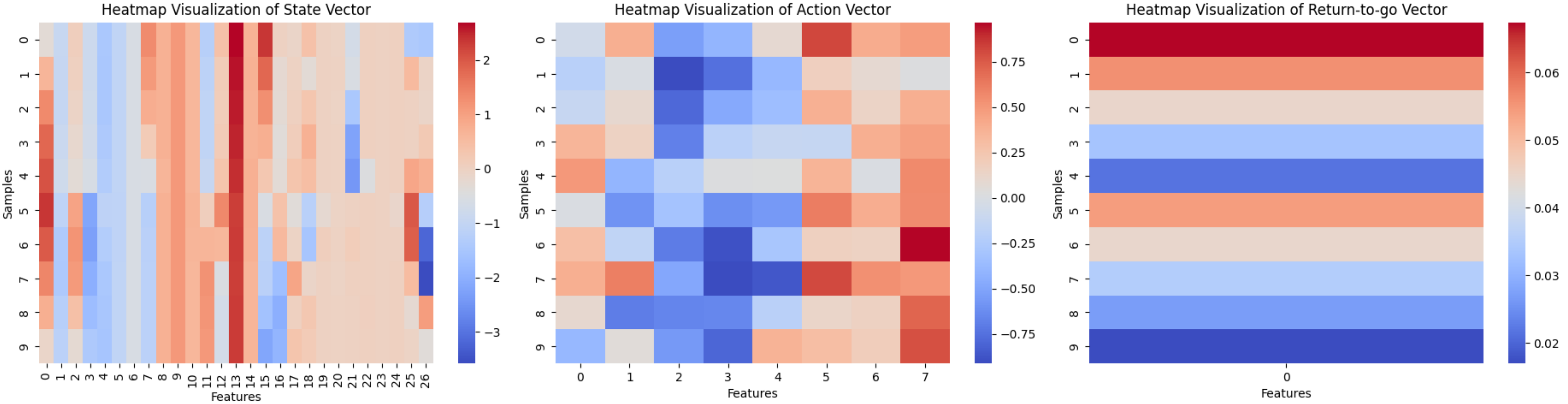}
    \caption{
    Ablation on the visualization of prompts.
    For the visualization, we generate state vectors, action vectors, and return-to-go vectors from 10 different samples. Samples 0-4 represent the initial prompts, while samples 5-9 represent the prompts after fine-tuning.
    The results provide valuable insights into the dimensions that significantly influence the final performance. }
    \label{fig:vis}
\end{figure}

We also generate heat maps to compare the original prompt (samples 0-4) with the prompt after fine-tuning (samples 5-9) in Figure \ref{fig:vis}. 
The heat maps illustrate significant differences in certain dimensions resulting from prompt-tuning. 
Specifically, when examining the State Vector, we observe a consistent trend of low values in dimensions such as the 7th and 15th. This pattern is also observed in the Action Vector and Return-to-go Vector. 
While Figure \ref{fig:vis} itself may not provide directly interpretable content, it offers insights into the dimensions that play a critical role in determining the final performance.

\newpage
\section{Detailed Results}
\label{sec:apdr}
Table \ref{tab:baseline} presents the normalized scores used for normalization, as proposed in \cite{hafner2020mastering, d4rl}. 
Table \ref{tab:raw_d4rl} shows the raw scores corresponding to Tables \ref{tab:d4rl}.

\begin{table}[h]
\caption{Baseline scores used for normalization.}
 \label{tab:baseline}
 \vskip 0.15in
\centering
\small
\begin{tabular}{lrr}
\toprule
\multicolumn{1}{c}{\bf Game} & \multicolumn{1}{c}{\bf Random} & \multicolumn{1}{c}{\bf Gamer} \\
  \midrule
HalfCheet & $-280.2$ & $12135$ \\
Hopper & $-20.3$ & $3234.3$ \\
Walker & $1.6$ &	$4592.3$ \\
 \bottomrule
 \end{tabular}
\end{table}

\begin{table*}[h]
\centering
\caption{Raw results for D4RL with medium datasets. 
%
The best mean scores are highlighted in bold.
Each environment has prompts of length $K^*=5$ and is fine-tuned at the limited expert dataset with three different seeds.
The results reveal that both PTDT-offline and PTDT-online methods exhibit similar performance, and they both achieve comparable results to full-model fine-tuning approaches. 
%
}
\label{tab:raw_d4rl}
\vspace{.2cm}
\scalebox{0.8}{%
\begin{tabular}{>{\centering}p{0.14\textwidth}|>{\centering}p{0.14\textwidth}>{\centering}p{0.14\textwidth}>{\centering}p{0.14\textwidth}|>{\centering}p{0.18\textwidth}>{\centering}p{0.14\textwidth}>{\centering\arraybackslash}p{0.14\textwidth}}
\toprule
Environment & CQL &   DT & Prompt-DT & Prompt-DT-FT  & PTDT-offline & PTDT-online \\ \midrule
HalfCheetah  & \textbf{5232.2} & 5008.7 & 4996.3  & 5021.1 &  5021.1 &  5008.7 \\
Hopper       & 1867.4 &   2179.8 & 2222.1  & \textbf{2293.7} & 2270.9  & 2248.2 \\
Walker       & \textbf{3637.4} &  3398.7 & 3426.3  & 3476.8 &  3504.3 & 3499.7 \\ \bottomrule
\end{tabular}%
}
\end{table*}


\end{document}